\pdfoutput=1

\documentclass[11pt]{article}

\usepackage{EMNLP2023}

\usepackage{times}
\usepackage{latexsym}

\usepackage[T1]{fontenc}

\usepackage[utf8]{inputenc}

\usepackage{microtype}

\usepackage{inconsolata}
\usepackage{booktabs}
\usepackage{graphicx}
\usepackage{amsfonts} 
\usepackage{amsmath}
\DeclareUnicodeCharacter{2212}{-}

\title{Towards Probing Contact Center Large Language Models} 

\author{Varun Nathan, Ayush Kumar, Digvijay Ingle \and Jithendra Vepa \\
        \texttt{\{varun.nathan, ayush, digvijay.ingle jithendra\}@observe.ai}\\
        Observe.AI \\ Bangalore, India}

\begin{document}
\maketitle
\begin{abstract}
Fine-tuning large language models (LLMs) with domain-specific instructions has emerged as an effective method to enhance their domain-specific understanding. Yet, there is limited work that examines the core characteristics acquired during this process. In this study, we benchmark the fundamental characteristics learned by contact-center (CC) specific instruction fine-tuned LLMs with out-of-the-box (OOB) LLMs via probing tasks encompassing conversational, channel, and automatic speech recognition (ASR) properties. We explore different LLM architectures (Flan-T5 and Llama), sizes (3B, 7B, 11B, 13B), and fine-tuning paradigms (full fine-tuning vs PEFT). Our findings reveal remarkable effectiveness of CC-LLMs on the in-domain downstream tasks, with improvement in response acceptability by over 48\% compared to OOB-LLMs. Additionally, we compare the performance of OOB-LLMs and CC-LLMs on the widely used SentEval dataset, and  assess their capabilities in terms of surface, syntactic, and semantic information through probing tasks.  Intriguingly, we note a relatively consistent performance of probing classifiers on the set of probing tasks. Our observations indicate that CC-LLMs, while outperforming their out-of-the-box counterparts, exhibit a tendency to rely less on encoding surface, syntactic, and semantic properties, highlighting the intricate interplay between domain-specific adaptation and probing task performance opening up opportunities to explore behavior of fine-tuned language models in specialized contexts.

\end{abstract}

\section{Introduction}
\label{section:introduction}







Language models (LMs) have made significant strides in recent years, with their ability to generate coherent and contextually relevant text garnering attention from researchers and practitioners alike \cite{DBLP:conf/iclr/WeiBZGYLDDL22, DBLP:journals/corr/abs-2303-08774}. These models, trained on massive amounts of data, have demonstrated their proficiency across a range of natural language processing tasks, including machine translation, sentiment analysis, and text summarization. Researchers have also explored the potential of fine-tuning these general-purpose LMs on domain-specific data, leading to improved performance in areas such as biomedical research \cite{DBLP:journals/bib/LuoSXQZPL22, singhal2023large}, coding \cite{DBLP:conf/emnlp/0034WJH21}, and finance \cite{DBLP:journals/corr/abs-2303-17564}. However, one domain that has received relatively little attention is the contact center industry. Contact centers play a crucial role in customer service and support for various businesses. They handle a wide range of customer queries, ranging from technical support to billing inquiries. The effectiveness of these interactions directly impacts customer satisfaction and, ultimately, business success. Integrating LMs into contact center operations has the potential to revolutionize the industry. For example, contact center agents can leverage LMs to access a vast array of information and generate personalized, contextually appropriate responses in real-time. However, conversations in contact centers often involve domain-specific knowledge, jargon, and abbreviations, posing challenges for traditional LMs to comprehend accurately. Moreover, the unique conversational dynamics and customer service etiquettes in contact centers further complicate the task of capturing domain-specific nuances effectively. 

Instruction fine-tuning \cite{DBLP:conf/icml/LongpreHVWCTZLZ23} has emerged as one the promising approaches to develop domain-specific LMs generalizable to numerous tasks. However, the effectiveness and applicability of this technique in the contact center domain have not been thoroughly investigated.

In recent years, parameter-efficient methods have gained popularity for fine-tuning large language models (LLMs) efficiently with limited computational resources while preserving performance. Among these techniques, Low-Rank Adaptation (LoRA) \cite{DBLP:journals/corr/abs-2106-09685} has gained traction due to its advantages in training speed and inference latency, often outperforming full fine-tuning. However, its suitability and effectiveness in the contact center domain remains largely unexplored.


In this paper, we explore the potential of instruction fine-tuning in improving language model performance in contact-center domain and seek to address the following research questions:

\begin{itemize}
    \item \textbf{RQ1:} How effective is instruction fine-tuning in improving the performance of LLMs on downstream tasks in contact-center domain?

    \item \textbf{RQ2:} What specific properties unique to contact-center (CC) interactions are acquired by LLMs fine-tuned on CC instruction sets in contrast to out-of-the-box (OOB) models?

    \item \textbf{RQ3:} How does the choice of model architecture and size shape the performance of LLMs on probing tasks?

    \item  \textbf{RQ4:} How do the fundamental characteristics learned when fine-tuning LLMs with parameter-efficient methods differ compared to traditional full fine-tuning methods?

    \item \textbf{RQ5:} Once fine-tuned on a domain-specific instruction, what general purpose fundamental properties do LLMs retain?
\end{itemize}

To address these questions, we conduct a comprehensive analysis of CC-LLMs examining the linguistic patterns, domain-specific knowledge, and conversational dynamics that influence the LLM's performance. By shedding light on the unique properties of CC interactions and investigating the potential of instruction fine-tuning, this research aims to contribute to the advancement of language models in specialized domains. Ultimately, our findings can pave the way for more effective and efficient customer interactions in contact centers, benefiting both service providers and customers alike.

\section{Training Contact-Center Instruction-Tuned LM}
\label{section:training-llm}





Numerous closed-source \cite{DBLP:conf/nips/BrownMRSKDNSSAA20, DBLP:journals/corr/abs-2303-08774} and open-source \cite{DBLP:journals/corr/abs-2307-09288} general purpose LLMs have demonstrated abilities to address a diverse range of tasks in natural language processing. However, specialised models like CodeT5 \cite{DBLP:conf/emnlp/0034WJH21}, StarCoder \cite{DBLP:journals/corr/abs-2305-06161}, Med-PaLM \cite{singhal2023large}, BioGPT \cite{DBLP:journals/bib/LuoSXQZPL22}, Galactica \cite{DBLP:journals/corr/abs-2211-09085}, BloombergGPT \cite{DBLP:journals/corr/abs-2303-17564} emphasize the significance of domain-specific models in achieving exceptional performance within fields like coding, bio-medicine, science, and finance. These models excel at producing high-quality outputs and tackling domain-specific challenges, illustrating the need of tailored LMs in diverse domains.


Inspired by the above works, we leverage in-house dataset\footnote{We cannot release the dataset due to proprietary reasons.} of conversational interactions between agents and customers to train a CC-specific LLM (CC-LLM) to model the properties of CC conversations. Due to the spontaneous nature of these conversations, the data is often nuanced with characteristics such as multi-party speakers, disfluencies, overtalks, call transfers, etc. Furthermore, the data is obtained post transcription from an automatic speech recognition (ASR) system, thus introducing the challenge of dealing with ASR errors such as insertions, deletions, and substitutions, in turn establishing the need for a model robust to the conversational properties. In this work, we adopt an approach of instruction fine-tuning \cite{DBLP:conf/iclr/WeiBZGYLDDL22, DBLP:conf/icml/LongpreHVWCTZLZ23}, which is fine-tuning the language model on a mixture of tasks expressed via natural language instructions. 

The process of fine-tuning a LM for contact-center applications involves three main components: 
a contact-center dataset, instructions specific to contact center use-cases, and a language model. To curate the contact-center dataset, we collect ASR transcripts of English conversations between agents and customers from various sectors, such as e-commerce, ed-tech, logistics, etc. We observe an average word-error-rate (WER) of 14.3 on these transcripts. The next step is to gather the instructions and their corresponding responses from the collected calls. We employ three processes to obtain these instructions: 
\begin{itemize}
    \item  Initially, we utilize our previously annotated data from use-cases such as sentiment detection, intent classification, entity recognition, and question answering. We reformat this data into triplets containing an instruction, input, and output. The instructions and outputs for these tasks are aggregated through a semi-automatic process involving human intervention. We leverage the human-in-the-loop approach to generate instructions and their variations that can elicit the desired response for the given task.

    \item  Following this, we expand the instructions by employing a paraphrasing process. This allows us to generate multiple styles of the same instructions, thereby increasing the diversity of the instruction set.

    \item  In addition to using the annotated data from the past, we also gather new sets of instructions by instructing human annotators to generate relevant questions that can be asked and answered during a call. Similar to the previous step, we expand these generated instructions using the paraphrasing process.
\end{itemize}

To assist the annotators in generating these tasks, we provide them with a list of insights that we aim to extract from the calls to address various use-cases. Examples of such insights include understanding and tracking customer and agent behaviors, following the steps taken in the call to resolve customer issues, and identifying different objections raised by the customers.

Here are some important statistics on the internally curated contact-center dataset:
\begin{itemize}
    \item Total corpus size: 110030
    \item Number of instructions: 2468
    \item Number of tasks: 59
\end{itemize}
Some example tasks considered in the dataset include \textit{reason for call},  \textit{call summarization}, \textit{segmented call summarization}, \textit{confirmed next steps}, \textit{Question-Answering (QA)}, \textit{entity extraction}, \textit{topic segmentation} and \textit{text rewriting}. Refer to Appendix \ref{appendix:task-definitions} for definition of these tasks.

Further, we fine-tune OOB-LLMs that are free for commercial use on the curated dataset. Specifically, we obtain CC-Flan-T5 model by fine-tuning the corresponding sized OOB-Flan-T5 model, and obtain CC-Llama model by fine-tuning the corresponding sized OOB-Llama-Instruct model.




\section{Probing tasks}
\label{section:probing-tasks}






In this section, we delve into the probing tasks employed to uncover the various properties learned by LMs, with a specific focus on the characteristics that are fundamental to effectively understand the context of CC interactions. Probing tasks tailored to the CC domain provide valuable insights into the capabilities and limitations of LMs in this specific area, as demonstrated in a previous study \cite{DBLP:conf/blackboxnlp/KumarSV21}. In their work, the authors propose probing tasks to investigate the conversational, channel, and ASR properties of pre-trained LMs. Given our own work in the contact-center domain, we refer to these probing tasks and utilize the details outlined in the work to construct datasets \footnote{We cannot release the dataset due to proprietary reasons.} for a set of classifier-based probing tasks.

Additionally, we also probe the LMs on a benchmark probing task of SentEval suite \cite{DBLP:conf/acl/BaroniBLKC18} that aims to uncover the linguistic knowledge and underlying properties learned by the model. SentEval suite consists of probing tasks across the categories of surface information, syntactic information and semantic information.


\section{Experiment Design}






To address the research questions outlined in Section \ref{section:introduction}, we design a series of experiments investigating the impact of model size, architecture, fine-tuning paradigms and evaluate the properties learned by language models in the CC domain.

Firstly, we compare three types of models: OOB foundation model, OOB instruction model, and the CC instruction model. At first, we compare these models on the CC-specific downstream tasks (RQ1) to understand the quality of generated responses given the call-transcript and an instruction. 

Secondly, we study the differences in performance across these models in terms of their capability to exhibit the learning of fundamental characteristics of CC data (RQ2). We delve into the impact of model size and architecture (RQ3). We compare different LLMs to explore how the choice of model architecture and size influences their performance on probing tasks. This investigation is critical in unraveling the intricate relationship between model design choices and the underlying properties learned by LLMs.

Thirdly, as fine-tuning is a crucial aspect of training LLMs, we also compare two fine-tuning paradigms: full fine-tuning and LoRA-based parameter-efficient fine-tuning (PEFT). By contrasting these approaches, we aim to uncover how the fundamental characteristics learned by LLMs differ under each method. This investigation is directly tied to RQ4, where we seek to understand the differences between traditional full fine-tuning and PEFT methods.

Finally, to evaluate the general-purpose properties represented by surface, syntactic and semantic information learnt by LLMs, we utilize the SentEval dataset (RQ5). For probing the dataset, we train a one-layer linear MLP classifier, following the previous work by \citealp{DBLP:conf/iclr/AlainB17}.

\begin{figure*}[t]
  \centering
  \includegraphics[width=0.8\textwidth]{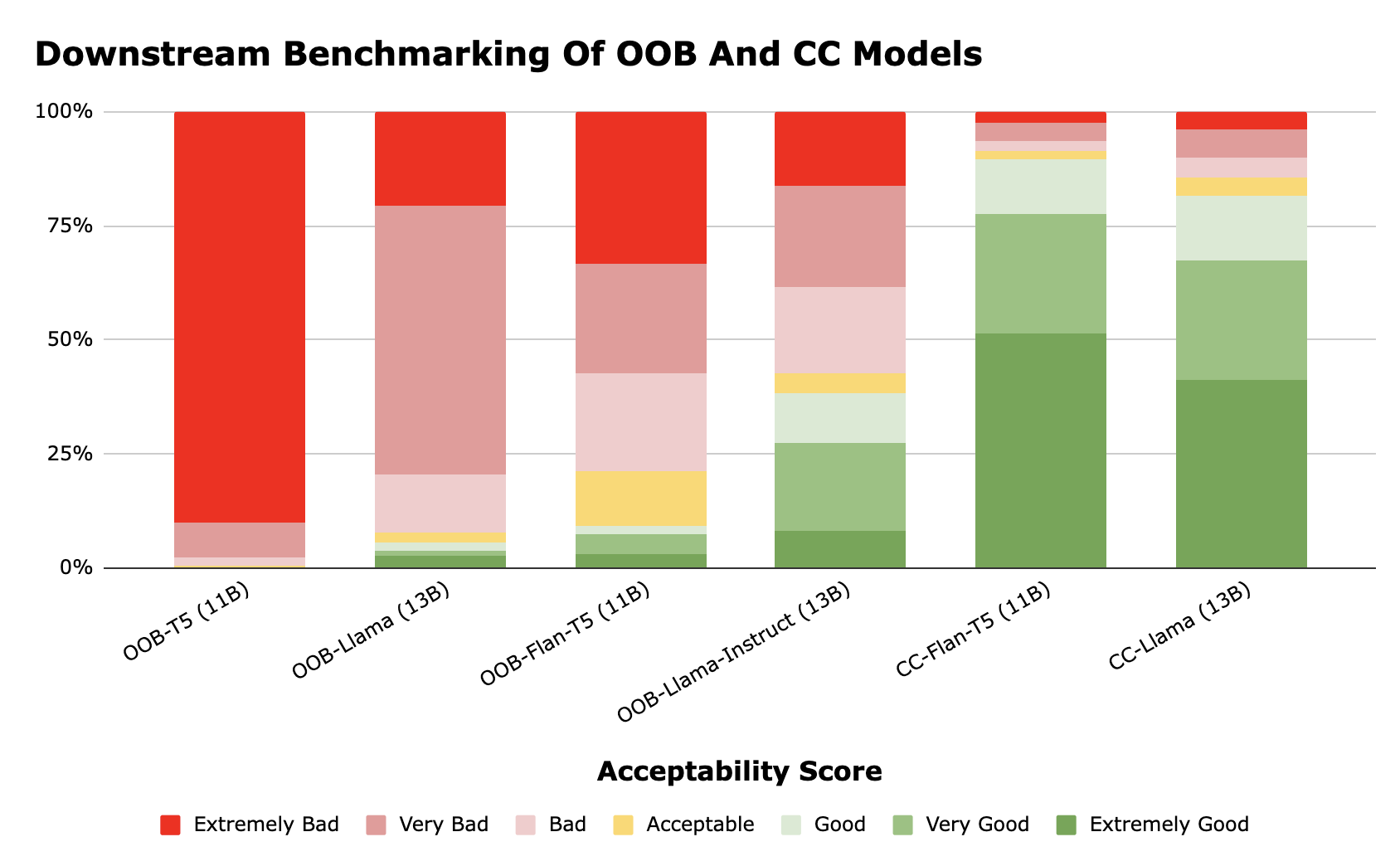}
  \caption{Benchmarking quality of responses generated by CC LLMs versus OOB LLMs on downstream tasks in contact-center domain using a 7-point scale of \textit{Extremely Bad} response to \textit{Extremely Good} response. We note that CC LLMs result in over 48\% improvement in response acceptability compared to OOB LLMs.}
  \label{fig:oob-vs-cc-models}
\end{figure*}

\begin{table*}[t]
    \footnotesize
    \setlength{\tabcolsep}{4pt}
    \begin{tabular}{lcccc|cccc|cccc}
    \toprule
     &
      \multicolumn{4}{c}{\textbf{OOB Foundation}} &
      \multicolumn{4}{|c|}{\textbf{OOB Instruction-Tuned}} &
      \multicolumn{4}{c}{\textbf{Contact Center}} \\
    \midrule
    \textbf{\begin{tabular}[c]{@{}l@{}}Probing \\ Tasks\end{tabular}} &
      \multicolumn{1}{l}{\textbf{\begin{tabular}[c]{@{}l@{}}OOB-\\T5\\ (3B)\end{tabular}}} &
      \multicolumn{1}{l}{\textbf{\begin{tabular}[c]{@{}l@{}}OOB-\\T5\\ (11B)\end{tabular}}} &
      \multicolumn{1}{l}{\textbf{\begin{tabular}[c]{@{}l@{}}OOB-\\ Llama\\ (7B)\end{tabular}}} &
      \multicolumn{1}{l|}{\textbf{\begin{tabular}[c]{@{}l@{}}OOB-\\ Llama\\ (13B)\end{tabular}}} &
      \multicolumn{1}{l}{\textbf{\begin{tabular}[c]{@{}l@{}}OOB-\\ Flan-\\ T5\\(3B)\end{tabular}}} &
      \multicolumn{1}{l}{\textbf{\begin{tabular}[c]{@{}l@{}}OOB-\\ Flan-\\ T5\\(11B)\end{tabular}}} &
      \multicolumn{1}{l}{\textbf{\begin{tabular}[c]{@{}l@{}}OOB-\\ Llama-\\ Instruct\\ (7B)\end{tabular}}} &
      \multicolumn{1}{l|}{\textbf{\begin{tabular}[c]{@{}l@{}}OOB-\\ Llama-\\ Instruct\\ (13B)\end{tabular}}} &
      \multicolumn{1}{l}{\textbf{\begin{tabular}[c]{@{}l@{}}CC-\\ Flan-\\ T5\\(3B)\end{tabular}}} &
      \multicolumn{1}{l}{\textbf{\begin{tabular}[c]{@{}l@{}}CC-\\ Flan-\\ T5\\(11B)\end{tabular}}} &
      \multicolumn{1}{l}{\textbf{\begin{tabular}[c]{@{}l@{}}CC-\\ Llama-\\ (7B)\end{tabular}}} &
      \multicolumn{1}{l}{\textbf{\begin{tabular}[c]{@{}l@{}}CC-\\ Llama-\\ (13B)\end{tabular}}} \\
    \midrule
    Disfluency &
      72.12 &
      71.97 &
      68.30 &
      71.57 &
      71.72 &
      73.03 &
      68.88 &
      69.81 &
      72.24 &
      72.89 &
      69.16 &
      67.83 \\
    Pause &
      80.90 &
      80.70 &
      77.79 &
      81.25 &
      82.09 &
      83.45 &
      80.45 &
      80.25 &
      81.78 &
      83.00 &
      76.85 &
      79.24 \\
    Overtalk &
      86.95 &
      89.55 &
      82.79 &
      81.59 &
      89.45 &
      90.70 &
      83.25 &
      77.70 &
      87.80 &
      88.19 &
      72.55 &
      78.92 \\
    Question &
      77.52 &
      74.49 &
      70.34 &
      74.31 &
      77.59 &
      75.39 &
      73.03 &
      74.33 &
      76.96 &
      80.37 &
      76.22 &
      77.15 \\
    Speaker &
      80.95 &
      81.96 &
      77.54 &
      82.70 &
      82.55 &
      83.39 &
      80.26 &
      80.21 &
      82.11 &
      82.72 &
      78.70 &
      79.94 \\
    Turn Length &
      67.65 &
      69.35 &
      66.23 &
      69.09 &
      69.20 &
      69.66 &
      66.03 &
      67.29 &
      68.88 &
      68.79 &
      67.27 &
      67.95 \\
    Turn Taking &
      68.30 &
      69.14 &
      65.01 &
      69.33 &
      64.30 &
      67.66 &
      69.62 &
      68.65 &
      66.83 &
      69.59 &
      62.50 &
      63.45 \\
    \begin{tabular}[c]{@{}l@{}}Token \\ Multi-class\end{tabular} &
      52.45 &
      49.32 &
      40.71 &
      42.64 &
      59.91 &
      63.07 &
      43.02 &
      40.60 &
      59.31 &
      60.73 &
      41.62 &
      42.85 \\
     Token Binary &
      60.50 &
      60.48 &
      50.07 &
      54.93 &
      68.34 &
      73.12 &
      49.84 &
      48.77 &
      70.11 &
      70.07 &
      49.88 &
      50.14 \\
    \midrule
    \begin{tabular}[c]{@{}l@{}}Average \\ Score\end{tabular}  &
      71.93 &
      71.88 &
      66.53 &
      69.71 &
      73.90 &
      75.50 &
      68.26 &
      67.51 &
      74.00 &
      75.15 &
      66.08 &
      67.50 \\
    \bottomrule
    \end{tabular}
    \caption{Benchmarking CC and OOB LLMs in terms of Macro F1 evaluated on conversational probing tasks. We note a mixed trend where 2 out of 4 CC LMs exhibit higher or comparable average score than OOB Foundation LMs and OOB Instruction-tuned LMs.}
    \label{tab:internal-probing-results}
\end{table*}

\begin{table*}[t]
    \footnotesize
    \centering
    \begin{tabular}{lcc|cc}
    \toprule
     &
      \multicolumn{2}{c|}{\textbf{Encoder-Decoder}} &
      \multicolumn{2}{c}{\textbf{Decoder Only}} \\
    \midrule
    \textbf{\begin{tabular}[c]{@{}l@{}}Probing \\ Tasks\end{tabular}} &
      \multicolumn{1}{l}{\textbf{\begin{tabular}[c]{@{}l@{}}CC-Flan-\\ T5\\(11B)\end{tabular}}} &
      \multicolumn{1}{l|}{\textbf{\begin{tabular}[c]{@{}l@{}}CC-Flan-\\ T5-PEFT \\ (11B)\end{tabular}}} &
      \multicolumn{1}{l}{\textbf{\begin{tabular}[c]{@{}l@{}}CC-Llama \\ (13B)\end{tabular}}} &
      \multicolumn{1}{l}{\textbf{\begin{tabular}[c]{@{}l@{}}CC-Llama-\\ PEFT\\ (13B)\end{tabular}}} \\
    \midrule
    Disfluency &
      72.89 &
      60.77 &
      67.83 &
      71.01 \\
    Pause &
      83.00 &
      82.13 &
      79.24 &
      73.04 \\
    Overtalk &
      88.19 &
      77.12 &
      78.92 &
      68.94 \\
    Question &
      80.37 &
      76.76 &
      77.15 &
      57.72 \\
    Speaker &
      82.72 &
      81.57 &
      79.94 &
      77.40 \\
    Turn Length &
      68.79 &
      68.30 &
      67.95 &
      66.41 \\
    Turn Taking &
      69.59 &
      67.22 &
      63.45 &
      57.07 \\
    \begin{tabular}[c]{@{}l@{}}Token \\ Multi-class\end{tabular} &
      60.73 &
      56.27 &
      42.85 &
      41.26 \\
    Token Binary &
      70.07 &
      65.44 &
      50.14 &
      36.29 \\
    \midrule
    Average Score &
      \multicolumn{1}{c}{77.94} &
      \multicolumn{1}{c}{73.41} &
      \multicolumn{1}{c}{73.50} &
      \multicolumn{1}{c}{67.37} \\
    \bottomrule
    \end{tabular}
    \caption{Benchmarking conversational properties acquired by CC LLMs when fine-tuned using various paradigms (full-tuning and PEFT). While reducing parameters and memory requirements, PEFT results in a trade-off with probing-task performance, calling for careful consideration when adopting PEFT for LLM fine-tuning.} 
    \label{tab:peft-benchmarks}
\end{table*}

\begin{table*}[t]
    \footnotesize
    \begin{tabular}{lcc|cc|cc}
    \toprule
           & \multicolumn{2}{c}{\textbf{OOB Foundation}} & \multicolumn{2}{|c|}{\textbf{OOB Instruction Tuned}} & \multicolumn{2}{c}{\textbf{Contact Center}} \\
    \midrule
        \textbf{Probing Tasks} &
          \textbf{\begin{tabular}[c]{@{}c@{}}OOB-T5\\ (11B)\end{tabular}} &
          \textbf{\begin{tabular}[c]{@{}c@{}}OOB-Llama\\ (13B)\end{tabular}} &
          \textbf{\begin{tabular}[c]{@{}c@{}}OOB-Flan-T5\\ (11B)\end{tabular}} &
          \textbf{\begin{tabular}[c]{@{}c@{}}OOB-Llama\\ -Instruct (13B)\end{tabular}} &
          \textbf{\begin{tabular}[c]{@{}c@{}}CC-Flan-T5\\ (11B)\end{tabular}} &
          \textbf{\begin{tabular}[c]{@{}c@{}}CC-Llama\\ (13B)\end{tabular}} \\
    \midrule
        Bigram Shift           & 92.48\%              & 85.66\%              & 94.19\%                  & 85.59\%                 & 92.17\%              & 76.79\%              \\
        Coordination Inversion & 79.36\%              & 68.65\%              & 77.59\%                  & 71.68\%                 & 76.59\%              & 70.30\%              \\
        Object Number          & 82.70\%              & 73.90\%              & 89.20\%                  & 74.20\%                 & 86.90\%              & 76.49\%              \\
        Odd Man Out            & 73.69\%              & 66.09\%              & 74.99\%                  & 66.90\%                 & 72.99\%              & 63.51\%              \\
        Past Present           & 88.99\%              & 84.17\%              & 89.19\%                  & 85.19\%                 & 89.59\%              & 82.98\%              \\
        Sentence Length        & 100.00\%             & 100.00\%             & 100.00\%                 & 100.00\%                & 100.00\%             & 100.00\%             \\
        Subj Number            & 86.19\%              & 79.49\%              & 92.09\%                  & 79.66\%                 & 90.29\%              & 81.57\%              \\
        Top Constituents       & 68.85\%              & 73.98\%              & 74.65\%                  & 67.44\%                 & 75.78\%              & 58.55\%              \\
        Tree Depth             & 36.02\%              & 28.73\%              & 37.24\%                  & 32.65\%                 & 38.49\%              & 27.92\%              \\
    \midrule
        \textbf{Average Score} & 78.70\%              & 73.41\%              & 81.02\%                  & 73.70\%                 & 80.31\%              & 70.90\%   \\       
    \bottomrule
    \end{tabular}
    \caption{Benchmarking CC and OOB LLMs in terms of Macro F1 evaluated on SentEval probing tasks. CC-Flan-T5 (11B) closely aligns with OOB-Flan-T5 (11B), demonstrating linguistic proficiency alongside its conversational capabilities, while CC-Llama 13B exhibits a lower performance, potentially attributed to its decoder-only architecture, prompting avenues for future exploration.}
    \label{tab:senteval_benchmarks}
\end{table*}

\section{Implementation Details}
\label{section:implementation-details}
In this section, we provide a detailed account of the implementation specifics related to our investigation into LLMs fine-tuned on CC instructions.

To initiate the process, we extract representations from the LLMs, harnessing their hidden states to encapsulate the contextual nuances present in the transcripts as well as instructions which are indicative of the tasks they are expected to perform as demonstrated in a previous study \cite{DBLP:journals/corr/abs-2305-14171}. Our approach is different from the authors in the sense that we use a linear probe as opposed to an attentional probe which is explained in more detail later in this section. For encoder-decoder models, we tap into the final encoder layer to obtain representations for each token within the input prompt. We adopt a suitable aggregation method depending on the characteristics of the specific probing task. For single-token probing tasks, we use the representation of the target token. For other tasks, we obtain an average of representations of all input tokens. On the other hand, in decoder-only models, we utilize the last hidden layer of the decoder block. The aggregation approach for decoder-only models aligns with encoder-decoder models for single-token probing tasks but relies on the last token's representation for other tasks. This difference stems from encoder-decoder models being bidirectional, making each token representation contextual to the entire sequence. In contrast, decoder models process tokens sequentially from left to right, making each token's representation contextual only to the tokens before it. Therefore, we consider the last token's representation as it encompasses information from entire sequence.

For encoder-decoder models, the embedding dimension spans 512, 1024, 2048, and 4096 tokens, while for decoder-only models, it encompasses 32001 and 65024 tokens. The different embedding dimensions for the two classes of models stems from the difference in model architectures and context lengths employed during pre-training and fine-tuning. We employed a context length of 512 for all models when extracting representations due to the input prompts having a maximum sequence length of 507 tokens across probing tasks. All models receive an input consisting of a prompt, which is generated from the input dialog, and an instruction that defines the probing task being conducted.

Post representation extraction, we employ a Multilayer Perceptron (MLP) comprising a single hidden layer, utilizing the extracted representations as feature inputs for probing. We adopt a \textit{sigmoid} and \textit{softmax} activation function for binary and multi-class classification respectively. We perform a hyper-parameter sweep over the range - number of neurons in the hidden layer $\in \{50, 100, 150, 200\}$, learning rate $\in \{1e−3, 1e−2, 5e−2\}$, batch size $\in \{4, 8, 16, 32, 64\}$ and choose the best setting as evaluated on eval set. Additionally, we employ Adam optimizer with a dropout rate of 0.3, incorporate a weight decay of 0.00001, and set the maximum number of epochs to 20. Moreover, all experiments include early stopping and check-pointing for the best model.

Our experiments comprising representation extraction and probe classifier training were conducted on an AWS cloud instance, specifically, the p4d.24xlarge instance, equipped with eight GPUs, each boasting 40 GB of memory. The process of extracting representations is computationally intensive, chiefly because of the substantial embedding dimensionality. On average, a single run of the representation extraction job for decoder-only models of size 13 billion parameters demands 8-10 hours for completion, whereas the corresponding timeframe for encoder-decoder models of size 11 billion parameters is considerably shorter, ranging from 1-2 hours. In contrast, the probing models present a lighter computational load and general taking around 0.5 hours for completion.

Finally, we evaluate the probe models on a held out test set using macro F1 score.

\section{Results and Analysis}
In this section, we provide a comprehensive analysis of the performance evaluation results, shedding light on the key observation made during our study - the striking contrast in response quality between CC-LLMs and OOB-LLMs.

\subsection{RQ1}
We perform a qualitative assessment of the responses generated by CC and OOB-LLMs by categorizing the responses generated by each of them into one among following seven classes:  \textit{Extremely Good}, \textit{Very Good}, \textit{Good}, \textit{Acceptable}, \textit{Bad}, \textit{Very Bad}, and \textit{Extremely Bad}. The annotation process involved crafting of task-specific guidelines, covering aspects such as consistency, relevance, and fluency of the generated responses. Additionally, we provided annotated examples to elucidate the criteria for each quality level, ensuring a consistent understanding among annotators. To minimize potential bias, annotators were kept unaware of the model's identity. We further ensured data consistency and quality by conducting a cross-annotator review, which maintained inter-annotator disagreement at levels below 10\%. We further analyze the responses generated by both LLM groups, and observe significant drift in the distribution of responses among the seven classes (refer Figure \ref{fig:oob-vs-cc-models}). Specifically, responses generated by OOB-T5 (11B), OOB-Flan-T5 (11B), OOB-Llama (13B) and OOB-Llama-Instruct (13B) models are consistently skewed towards the lower end of the quality spectrum. A majority of these responses fell within the \textit{Bad} to \textit{Extremely Bad} categories, indicating that without specific fine-tuning, these models struggled to generate satisfactory responses for contact center-related instructions. Conversely, responses generated by CC-Flan-T5 (11B) and CC-Llama (13B) models exhibited a notable shift towards higher quality categories. A substantial portion of responses generated by these models landed in the \textit{Acceptable} to \textit{Extremely Good} range, demonstrating their ability to comprehend and generate contextually relevant responses for contact center interactions. We hypothesize that this disparity in performance can be attributed to the fine-tuning process with contact center data. It appears that by exposing the LLMs to domain-specific information and scenarios, they have acquired a deeper understanding of contact center interactions. 

\subsection{RQ2}
In order to investigate the conversational properties learnt by CC-LLMs that lead to performance superior to OOB-LLMs, we evaluate these models on the probing tasks in Section \ref{section:probing-tasks} and per the methodology described in Section \ref{section:implementation-details}. Although our probing tasks are carefully designed to uncover the latent knowledge within these models, our findings in Table \ref{tab:internal-probing-results} did not conclusively favor either type of LLM. Specifically, we observe a mixed trend where 1 out of 4 CC models, CC-Flan-T5 (3B) have higher average score and 2 out of 4 models, CC-Flan-T5 (11B) and CC-Llama (13B), have marginally lower ($<$ 0.5\%) average score compared to their corresponding OOB instruction-tuned counterparts. We also note a similar observation when comparing CC-LLMs with OOB foundation models wherein 3 out of 4 CC-LLMs have comparable or better average score. This intriguing result prompts us to delve deeper into several critical aspects of LLMs and their fine-tuning process prompting us to put forth following opportunities for exploration:

\begin{enumerate}
    \item  \textbf{Probing via Hidden Layer Representation:} While this method has been widely employed \cite{DBLP:conf/blackboxnlp/KumarSV21, DBLP:conf/blackboxnlp/FayyazAMMP21, DBLP:conf/blackboxnlp/ThukralKK21} to unearth linguistic properties by language models, we question whether it is sufficiently nuanced to capture conversational intricacies. It is conceivable that the differences we seek are not embedded in the representations themselves but are instead contingent on the decoding strategy employed during the language generation process. This insight underscores the pivotal role of decoding strategies in converting latent embeddings into coherent sequences of tokens that reflect both the given instruction and input. It prompts us to consider that instructing and fine-tuning a general-purpose model and a domain-specific model may ultimately hinge on decoding proficiency rather than vastly divergent learned representations. We believe that this calls for a deeper investigation into designing right probing strategies for recently popular generative language models trained via instruction fine-tuning.
    \item \textbf{Re-designing probing tasks:} Our existing set of probing tasks, although comprehensive, may not fully encapsulate the diverse landscape of conversational properties. Conversations are inherently dynamic, context-dependent, and influenced by various factors, including the interplay between participants, the history of the conversation, and the evolution of topics. Extracting hidden layer representations at a single utterance may not fully capture these dynamic aspects of conversation. It is plausible that more specific probing tasks tailored to the characteristic of contact center interactions are needed. These tasks should ideally mirror the challenges posed by real-world downstream applications that help diagnose the contextual properties and the interplay in the conversations.
\end{enumerate}

\subsection{RQ3}
From our results in Table \ref{tab:internal-probing-results}, we note that T5 models consistently outperform Llama models across the three settings, OOB Foundation, OOB Instruction-tuned and Contact Center, highlighting that T5's encoder-decoder architecture is better-equipped to comprehend conversational properties compared to Llama's decoder only architecture. Further, we observe that larger sizes generally translate to improved performance in both the OOB and CC settings reinforcing the pivotal role of model scale in grasping the complexities of conversation.

\subsection{RQ4}
Recently Parameter-Efficient Fine-Tuning, PEFT, has turned out to be a compelling approach to fine-tune large-scale pre-trained LMs while mitigating the challenges associated with resource-intensive full fine-tuning. Hence, we investigate the similarities and differences in the linguistic properties learnt by models fine-tuned using PEFT vs those fine-tuned in vanilla fashion (full fine-tuning). Specifically, we use Low-Rank Adaptation (LoRA) framework to perform parameter-efficient fine-tuning of OOB-Flan-T5 (11B) and OOB-Llama-Instruct (13B) models, using the instruction dataset described in Section \ref{section:training-llm} to obtain CC-Flan-T5-PEFT (11B) and CC-Llama-PEFT (13B) models respectively. We further probe CC-Flan-T5 (11B), CC-Flan-T5-PEFT (11B), CC-Llama (13B) and CC-Llama-PEFT (13B) on the probing tasks in Section \ref{section:probing-tasks}. Based on our results in Table \ref{tab:peft-benchmarks}, we observe that CC-Flan-T5-PEFT (11B) leads to a 4.53\% lower average score on the probing tasks compared to CC-Flan-T5 (11B). A similar observation also holds true for Llama models, where CC-Llama-PEFT (13B) results in a 6.13\% lower average score compared to CC-Llama (13B). This notable trend highlights that while the reduction in trainable parameters and memory requirements offers undeniable advantages in resource efficiency, it comes at a trade-off in task-specific performance. Hence, it is imperative for practitioners and researchers to weigh these trade-offs carefully when considering the adoption of PEFT for fine-tuning LLMs.

\subsection{RQ5}
Contact center models, fine-tuned on contact-center instruction dataset, not only embrace conversational nuances but also retain fundamental linguistic properties inherent in OOB models. Our results in Table \ref{tab:senteval_benchmarks} exemplify this, wherein we observe that CC-Flan-T5 (11B) achieves an average score of 80.31\% on SentEval probing tasks, demonstrating a close alignment with OOB-Flan-T5 (11B) (81.02\%). Furthermore, it is noteworthy that CC-Flan-T5 (11B) exhibits a SentEval score 1.61\% higher than the OOB-T5 (11B) model, further underscoring its enhanced linguistic knowledge in addition to conversational capabilities in the context of contact center applications. Thus, CC-Flan-T5 (11B), while adapting to the intricacies of contact center conversations, maintains linguistic proficiency akin to its OOB counterparts. This implies that the model can effortlessly navigate the linguistic aspects of text, even when tailored to a specific domain. On the other hand, we note that CC-Llama (13B) exhibits a lower average on SentEval tasks compared to its OOB counterparts. One plausible explanation may revolve around the inherent characteristics of the Llama model, which is designed as a decoder-only architecture. While CC-Llama (13B) and OOB-Llama-Instruct (13B) model exhibit similar average score on conversational probing tasks (refer Table \ref{tab:internal-probing-results}), we hypothesize that its decoder-only architecture might introduce subtle variations in its linguistic representations compared to the encoder-decoder architectures like Flan. This distinction may result in a slight dip in performance on tasks that primarily assess linguistic properties while maintaining similar understanding of conversational properties. However, we leave this hypothesis for future exploration.

\section{Related Works}

In recent years, there have been significant advancements in the field of language modeling, with a particular focus on training domain-specific language models. One notable work in this area is Med-PaLM \cite{singhal2023large}, developed by researchers in the medical domain. Med-PaLM surpassed previous models in terms of performance on medical question answering tasks. CodeLLAMA \cite{DBLP:journals/corr/abs-2308-12950}, a prominent family of language models, specializes in code generation and infilling tasks, stemming from LLAMA2 to cater to software development and programming needs.

In the field of natural language processing, there have been numerous studies aimed at understanding the inner workings of language models. Probing tasks have been employed as a means to evaluate the fundamental properties encoded within the representations of these models. Baroni et al. \cite{DBLP:conf/acl/BaroniBLKC18} introduced a collection of probing tasks in the SentEval suite \cite{DBLP:conf/lrec/ConneauK18} to assess sentence embedding representations of language models. This work paved the way for subsequent studies, such as Tenney et al. \cite{DBLP:conf/acl/TenneyDP19} and Lin et al. \cite{DBLP:conf/blackboxnlp/LinTF19}, who performed layer-wise probing of BERT to uncover its semantic and hierarchical awareness.

Exploring the self-attention mechanism of language models has also provided insights into their inner workings. \citet{DBLP:conf/emnlp/KovalevaRRR19} and \citet{DBLP:conf/blackboxnlp/ClarkKLM19} delved into the patterns exhibited by individual self-attention heads in BERT, offering insights into their roles and functionalities.

While most studies focus on probing general language models, there are also investigations into domain-specific models and properties. \citet{DBLP:conf/blackboxnlp/KumarSV21} investigated the representations of language models in contact-center domain, revealing that LMs encode conversational and speaker-type properties to a large extent without external supervision, but lose the linguistic understanding of dependency relations. \citet{jin2019probing} probed biomedical language models and demonstrated their high effectiveness on biomedical named entity recognition (NER) and natural language inference (NLI) tasks. Our works falls into the line of training and investigating a domain specific language model.

\section{Conclusion}
Our study contributes to the growing body of research on fine-tuning LLMs with domain-specific instructions. In this work, we demonstrate that CC-LLMs, CC-Flan-T5 and CC-Llama, exhibit superior performance on downstream tasks within the contact center domain. This finding highlights the effectiveness of fine-tuning LLMs with domain-specific instructions in enhancing their understanding and applicability in specific domains. Furthermore, our comparison between OOB and CC models on the probing task reveals interesting observations. While the performance of probing classifiers on the set of probing tasks is relatively similar, indicating comparable contact-center specific properties encoding capabilities, the CC-LLMs still outperform OOB models. This suggests that the CC-LLMs possess additional domain-specific knowledge or contextual understanding that aids in achieving superior performance on downstream tasks. We also find that CC-LLMs, rely less on encoding surface, syntactic, and semantic properties. This indicates that these models may leverage other mechanisms or information sources to excel in the contact center domain thus opening opportunities for further exploration in this area.

\section*{Limitations}
While our study provides valuable insights into training a contact-center specific language model and conducting linear edge probing, it is important to acknowledge certain limitations in our work. Firstly, our exploration of language models is limited to a couple of models belonging to two architectures, one encoder-decoder and one decoder style. We choose these models on the basis of their effectiveness across different tasks as has been surfaced up in the research community, however, the trends we observe may not necessarily hold true for other models within the same class of architecture. Secondly, our work is based on the probing methodology of linear edge probing, which applies a one layer linear MLP on hidden representations. The performance and observations on probing tasks may differ if a different probing setup, such as an attention-based probing, is used. It is crucial to explore alternative probing methods to gain a more comprehensive understanding of the language model's characteristics. Moreover, the set of probing tasks we utilize may not cover the full range of characteristics that a language model can encode. Additional probing tasks can be considered to do a more extensive study of the model's capabilities. Lastly, our research is conducted on a proprietary dataset that cannot be released. This limits the ability of other researchers to directly compare their results or replicate our experiments. Access to the dataset is crucial for future work in this area, and we encourage the development of publicly available datasets for domain-specific language models.

Despite these limitations, our study underscores the importance of domain-specific instruction models and highlights the limited capacity of general-purpose language models to meet domain specific use-cases. Furthermore, we pose thought-provoking questions that can guide further research and contribute to the advancement of the research community's understanding of the properties encoded in generative language models in the new era.

\bibliography{anthology,custom}

\begin{thebibliography}{25}
\expandafter\ifx\csname natexlab\endcsname\relax\def\natexlab#1{#1}\fi

\bibitem[{Alain and Bengio(2017)}]{DBLP:conf/iclr/AlainB17}
Guillaume Alain and Yoshua Bengio. 2017.
\newblock \href {https://openreview.net/forum?id=HJ4-rAVtl} {Understanding
  intermediate layers using linear classifier probes}.
\newblock In \emph{5th International Conference on Learning Representations,
  {ICLR} 2017, Toulon, France, April 24-26, 2017, Workshop Track Proceedings}.
  OpenReview.net.

\bibitem[{Amini and Ciaramita(2023)}]{DBLP:journals/corr/abs-2305-14171}
Afra Amini and Massimiliano Ciaramita. 2023.
\newblock \href {https://doi.org/10.48550/arXiv.2305.14171} {Probing in
  context: Toward building robust classifiers via probing large language
  models}.
\newblock \emph{CoRR}, abs/2305.14171.

\bibitem[{Brown et~al.(2020)Brown, Mann, Ryder, Subbiah, Kaplan, Dhariwal,
  Neelakantan, Shyam, Sastry, Askell, Agarwal, Herbert{-}Voss, Krueger,
  Henighan, Child, Ramesh, Ziegler, Wu, Winter, Hesse, Chen, Sigler, Litwin,
  Gray, Chess, Clark, Berner, McCandlish, Radford, Sutskever, and
  Amodei}]{DBLP:conf/nips/BrownMRSKDNSSAA20}
Tom~B. Brown, Benjamin Mann, Nick Ryder, Melanie Subbiah, Jared Kaplan,
  Prafulla Dhariwal, Arvind Neelakantan, Pranav Shyam, Girish Sastry, Amanda
  Askell, Sandhini Agarwal, Ariel Herbert{-}Voss, Gretchen Krueger, Tom
  Henighan, Rewon Child, Aditya Ramesh, Daniel~M. Ziegler, Jeffrey Wu, Clemens
  Winter, Christopher Hesse, Mark Chen, Eric Sigler, Mateusz Litwin, Scott
  Gray, Benjamin Chess, Jack Clark, Christopher Berner, Sam McCandlish, Alec
  Radford, Ilya Sutskever, and Dario Amodei. 2020.
\newblock \href
  {https://proceedings.neurips.cc/paper/2020/hash/1457c0d6bfcb4967418bfb8ac142f64a-Abstract.html}
  {Language models are few-shot learners}.
\newblock In \emph{Advances in Neural Information Processing Systems 33: Annual
  Conference on Neural Information Processing Systems 2020, NeurIPS 2020,
  December 6-12, 2020, virtual}.

\bibitem[{Clark et~al.(2019)Clark, Khandelwal, Levy, and
  Manning}]{DBLP:conf/blackboxnlp/ClarkKLM19}
Kevin Clark, Urvashi Khandelwal, Omer Levy, and Christopher~D. Manning. 2019.
\newblock \href {https://doi.org/10.18653/v1/W19-4828} {What does {BERT} look
  at? an analysis of bert's attention}.
\newblock In \emph{Proceedings of the 2019 {ACL} Workshop BlackboxNLP:
  Analyzing and Interpreting Neural Networks for NLP, BlackboxNLP@ACL 2019,
  Florence, Italy, August 1, 2019}, pages 276--286. Association for
  Computational Linguistics.

\bibitem[{Conneau and Kiela(2018)}]{DBLP:conf/lrec/ConneauK18}
Alexis Conneau and Douwe Kiela. 2018.
\newblock \href
  {http://www.lrec-conf.org/proceedings/lrec2018/summaries/757.html} {Senteval:
  An evaluation toolkit for universal sentence representations}.
\newblock In \emph{Proceedings of the Eleventh International Conference on
  Language Resources and Evaluation, {LREC} 2018, Miyazaki, Japan, May 7-12,
  2018}. European Language Resources Association {(ELRA)}.

\bibitem[{Conneau et~al.(2018)Conneau, Kruszewski, Lample, Barrault, and
  Baroni}]{DBLP:conf/acl/BaroniBLKC18}
Alexis Conneau, Germ{\'{a}}n Kruszewski, Guillaume Lample, Lo{\"{\i}}c
  Barrault, and Marco Baroni. 2018.
\newblock \href {https://doi.org/10.18653/v1/P18-1198} {What you can cram into
  a single {\textbackslash}{\textdollar}{\&}!{\#}* vector: Probing sentence
  embeddings for linguistic properties}.
\newblock In \emph{Proceedings of the 56th Annual Meeting of the Association
  for Computational Linguistics, {ACL} 2018, Melbourne, Australia, July 15-20,
  2018, Volume 1: Long Papers}, pages 2126--2136. Association for Computational
  Linguistics.

\bibitem[{Fayyaz et~al.(2021)Fayyaz, Aghazadeh, Modarressi, Mohebbi, and
  Pilehvar}]{DBLP:conf/blackboxnlp/FayyazAMMP21}
Mohsen Fayyaz, Ehsan Aghazadeh, Ali Modarressi, Hosein Mohebbi, and
  Mohammad~Taher Pilehvar. 2021.
\newblock \href {https://doi.org/10.18653/v1/2021.blackboxnlp-1.29} {Not all
  models localize linguistic knowledge in the same place: {A} layer-wise
  probing on bertoids' representations}.
\newblock In \emph{Proceedings of the Fourth BlackboxNLP Workshop on Analyzing
  and Interpreting Neural Networks for NLP, BlackboxNLP@EMNLP 2021, Punta Cana,
  Dominican Republic, November 11, 2021}, pages 375--388. Association for
  Computational Linguistics.

\bibitem[{Hu et~al.(2021)Hu, Shen, Wallis, Allen{-}Zhu, Li, Wang, and
  Chen}]{DBLP:journals/corr/abs-2106-09685}
Edward~J. Hu, Yelong Shen, Phillip Wallis, Zeyuan Allen{-}Zhu, Yuanzhi Li,
  Shean Wang, and Weizhu Chen. 2021.
\newblock \href {http://arxiv.org/abs/2106.09685} {Lora: Low-rank adaptation of
  large language models}.
\newblock \emph{CoRR}, abs/2106.09685.

\bibitem[{Jin et~al.(2019)Jin, Dhingra, Cohen, and Lu}]{jin2019probing}
Qiao Jin, Bhuwan Dhingra, William~W Cohen, and Xinghua Lu. 2019.
\newblock Probing biomedical embeddings from language models.
\newblock \emph{NAACL HLT 2019}, page~82.

\bibitem[{Kovaleva et~al.(2019)Kovaleva, Romanov, Rogers, and
  Rumshisky}]{DBLP:conf/emnlp/KovalevaRRR19}
Olga Kovaleva, Alexey Romanov, Anna Rogers, and Anna Rumshisky. 2019.
\newblock \href {https://doi.org/10.18653/v1/D19-1445} {Revealing the dark
  secrets of {BERT}}.
\newblock In \emph{Proceedings of the 2019 Conference on Empirical Methods in
  Natural Language Processing and the 9th International Joint Conference on
  Natural Language Processing, {EMNLP-IJCNLP} 2019, Hong Kong, China, November
  3-7, 2019}, pages 4364--4373. Association for Computational Linguistics.

\bibitem[{Kumar et~al.(2021)Kumar, Sundararaman, and
  Vepa}]{DBLP:conf/blackboxnlp/KumarSV21}
Ayush Kumar, Mukuntha~Narayanan Sundararaman, and Jithendra Vepa. 2021.
\newblock \href {https://doi.org/10.18653/v1/2021.blackboxnlp-1.25} {What
  {BERT} based language model learns in spoken transcripts: An empirical
  study}.
\newblock In \emph{Proceedings of the Fourth BlackboxNLP Workshop on Analyzing
  and Interpreting Neural Networks for NLP, BlackboxNLP@EMNLP 2021, Punta Cana,
  Dominican Republic, November 11, 2021}, pages 322--336. Association for
  Computational Linguistics.

\bibitem[{Li et~al.(2023)Li, Allal, Zi, Muennighoff, Kocetkov, Mou, Marone,
  Akiki, Li, Chim, Liu, Zheltonozhskii, Zhuo, Wang, Dehaene, Davaadorj,
  Lamy{-}Poirier, Monteiro, Shliazhko, Gontier, Meade, Zebaze, Yee, Umapathi,
  Zhu, Lipkin, Oblokulov, Wang, V, Stillerman, Patel, Abulkhanov, Zocca, Dey,
  Zhang, Moustafa{-}Fahmy, Bhattacharyya, Yu, Singh, Luccioni, Villegas,
  Kunakov, Zhdanov, Romero, Lee, Timor, Ding, Schlesinger, Schoelkopf, Ebert,
  Dao, Mishra, Gu, Robinson, Anderson, Dolan{-}Gavitt, Contractor, Reddy,
  Fried, Bahdanau, Jernite, Ferrandis, Hughes, Wolf, Guha, von Werra, and
  de~Vries}]{DBLP:journals/corr/abs-2305-06161}
Raymond Li, Loubna~Ben Allal, Yangtian Zi, Niklas Muennighoff, Denis Kocetkov,
  Chenghao Mou, Marc Marone, Christopher Akiki, Jia Li, Jenny Chim, Qian Liu,
  Evgenii Zheltonozhskii, Terry~Yue Zhuo, Thomas Wang, Olivier Dehaene, Mishig
  Davaadorj, Joel Lamy{-}Poirier, Jo{\~{a}}o Monteiro, Oleh Shliazhko, Nicolas
  Gontier, Nicholas Meade, Armel Zebaze, Ming{-}Ho Yee, Logesh~Kumar Umapathi,
  Jian Zhu, Benjamin Lipkin, Muhtasham Oblokulov, Zhiruo Wang, Rudra~Murthy V,
  Jason Stillerman, Siva~Sankalp Patel, Dmitry Abulkhanov, Marco Zocca, Manan
  Dey, Zhihan Zhang, Nour Moustafa{-}Fahmy, Urvashi Bhattacharyya, Wenhao Yu,
  Swayam Singh, Sasha Luccioni, Paulo Villegas, Maxim Kunakov, Fedor Zhdanov,
  Manuel Romero, Tony Lee, Nadav Timor, Jennifer Ding, Claire Schlesinger,
  Hailey Schoelkopf, Jan Ebert, Tri Dao, Mayank Mishra, Alex Gu, Jennifer
  Robinson, Carolyn~Jane Anderson, Brendan Dolan{-}Gavitt, Danish Contractor,
  Siva Reddy, Daniel Fried, Dzmitry Bahdanau, Yacine Jernite,
  Carlos~Mu{\~{n}}oz Ferrandis, Sean Hughes, Thomas Wolf, Arjun Guha, Leandro
  von Werra, and Harm de~Vries. 2023.
\newblock \href {https://doi.org/10.48550/arXiv.2305.06161} {Starcoder: may the
  source be with you!}
\newblock \emph{CoRR}, abs/2305.06161.

\bibitem[{Lin et~al.(2019)Lin, Tan, and Frank}]{DBLP:conf/blackboxnlp/LinTF19}
Yongjie Lin, Yi~Chern Tan, and Robert Frank. 2019.
\newblock \href {https://doi.org/10.18653/v1/W19-4825} {Open sesame: Getting
  inside bert's linguistic knowledge}.
\newblock In \emph{Proceedings of the 2019 {ACL} Workshop BlackboxNLP:
  Analyzing and Interpreting Neural Networks for NLP, BlackboxNLP@ACL 2019,
  Florence, Italy, August 1, 2019}, pages 241--253. Association for
  Computational Linguistics.

\bibitem[{Longpre et~al.(2023)Longpre, Hou, Vu, Webson, Chung, Tay, Zhou, Le,
  Zoph, Wei, and Roberts}]{DBLP:conf/icml/LongpreHVWCTZLZ23}
Shayne Longpre, Le~Hou, Tu~Vu, Albert Webson, Hyung~Won Chung, Yi~Tay, Denny
  Zhou, Quoc~V. Le, Barret Zoph, Jason Wei, and Adam Roberts. 2023.
\newblock \href {https://proceedings.mlr.press/v202/longpre23a.html} {The flan
  collection: Designing data and methods for effective instruction tuning}.
\newblock In \emph{International Conference on Machine Learning, {ICML} 2023,
  23-29 July 2023, Honolulu, Hawaii, {USA}}, volume 202 of \emph{Proceedings of
  Machine Learning Research}, pages 22631--22648. {PMLR}.

\bibitem[{Luo et~al.(2022)Luo, Sun, Xia, Qin, Zhang, Poon, and
  Liu}]{DBLP:journals/bib/LuoSXQZPL22}
Renqian Luo, Liai Sun, Yingce Xia, Tao Qin, Sheng Zhang, Hoifung Poon, and
  Tie{-}Yan Liu. 2022.
\newblock \href {https://doi.org/10.1093/bib/bbac409} {Biogpt: generative
  pre-trained transformer for biomedical text generation and mining}.
\newblock \emph{Briefings Bioinform.}, 23(6).

\bibitem[{OpenAI(2023)}]{DBLP:journals/corr/abs-2303-08774}
OpenAI. 2023.
\newblock \href {https://doi.org/10.48550/arXiv.2303.08774} {{GPT-4} technical
  report}.
\newblock \emph{CoRR}, abs/2303.08774.

\bibitem[{Rozi{\`{e}}re et~al.(2023)Rozi{\`{e}}re, Gehring, Gloeckle, Sootla,
  Gat, Tan, Adi, Liu, Remez, Rapin, Kozhevnikov, Evtimov, Bitton, Bhatt,
  Canton{-}Ferrer, Grattafiori, Xiong, D{\'{e}}fossez, Copet, Azhar, Touvron,
  Martin, Usunier, Scialom, and Synnaeve}]{DBLP:journals/corr/abs-2308-12950}
Baptiste Rozi{\`{e}}re, Jonas Gehring, Fabian Gloeckle, Sten Sootla, Itai Gat,
  Xiaoqing~Ellen Tan, Yossi Adi, Jingyu Liu, Tal Remez, J{\'{e}}r{\'{e}}my
  Rapin, Artyom Kozhevnikov, Ivan Evtimov, Joanna Bitton, Manish Bhatt,
  Cristian Canton{-}Ferrer, Aaron Grattafiori, Wenhan Xiong, Alexandre
  D{\'{e}}fossez, Jade Copet, Faisal Azhar, Hugo Touvron, Louis Martin, Nicolas
  Usunier, Thomas Scialom, and Gabriel Synnaeve. 2023.
\newblock \href {https://doi.org/10.48550/arXiv.2308.12950} {Code llama: Open
  foundation models for code}.
\newblock \emph{CoRR}, abs/2308.12950.

\bibitem[{Singhal et~al.(2023)Singhal, Azizi, Tu, Mahdavi, Wei, Chung, Scales,
  Tanwani, Cole-Lewis, Pfohl et~al.}]{singhal2023large}
Karan Singhal, Shekoofeh Azizi, Tao Tu, S~Sara Mahdavi, Jason Wei, Hyung~Won
  Chung, Nathan Scales, Ajay Tanwani, Heather Cole-Lewis, Stephen Pfohl, et~al.
  2023.
\newblock Large language models encode clinical knowledge.
\newblock \emph{Nature}, pages 1--9.

\bibitem[{Taylor et~al.(2022)Taylor, Kardas, Cucurull, Scialom, Hartshorn,
  Saravia, Poulton, Kerkez, and Stojnic}]{DBLP:journals/corr/abs-2211-09085}
Ross Taylor, Marcin Kardas, Guillem Cucurull, Thomas Scialom, Anthony
  Hartshorn, Elvis Saravia, Andrew Poulton, Viktor Kerkez, and Robert Stojnic.
  2022.
\newblock \href {https://doi.org/10.48550/arXiv.2211.09085} {Galactica: {A}
  large language model for science}.
\newblock \emph{CoRR}, abs/2211.09085.

\bibitem[{Tenney et~al.(2019)Tenney, Das, and
  Pavlick}]{DBLP:conf/acl/TenneyDP19}
Ian Tenney, Dipanjan Das, and Ellie Pavlick. 2019.
\newblock \href {https://doi.org/10.18653/v1/p19-1452} {{BERT} rediscovers the
  classical {NLP} pipeline}.
\newblock In \emph{Proceedings of the 57th Conference of the Association for
  Computational Linguistics, {ACL} 2019, Florence, Italy, July 28- August 2,
  2019, Volume 1: Long Papers}, pages 4593--4601. Association for Computational
  Linguistics.

\bibitem[{Thukral et~al.(2021)Thukral, Kukreja, and
  Kavouras}]{DBLP:conf/blackboxnlp/ThukralKK21}
Shivin Thukral, Kunal Kukreja, and Christian Kavouras. 2021.
\newblock \href {https://doi.org/10.18653/v1/2021.blackboxnlp-1.31} {Probing
  language models for understanding of temporal expressions}.
\newblock In \emph{Proceedings of the Fourth BlackboxNLP Workshop on Analyzing
  and Interpreting Neural Networks for NLP, BlackboxNLP@EMNLP 2021, Punta Cana,
  Dominican Republic, November 11, 2021}, pages 396--406. Association for
  Computational Linguistics.

\bibitem[{Touvron et~al.(2023)Touvron, Martin, Stone, Albert, Almahairi,
  Babaei, Bashlykov, Batra, Bhargava, Bhosale, Bikel, Blecher, Canton{-}Ferrer,
  Chen, Cucurull, Esiobu, Fernandes, Fu, Fu, Fuller, Gao, Goswami, Goyal,
  Hartshorn, Hosseini, Hou, Inan, Kardas, Kerkez, Khabsa, Kloumann, Korenev,
  Koura, Lachaux, Lavril, Lee, Liskovich, Lu, Mao, Martinet, Mihaylov, Mishra,
  Molybog, Nie, Poulton, Reizenstein, Rungta, Saladi, Schelten, Silva, Smith,
  Subramanian, Tan, Tang, Taylor, Williams, Kuan, Xu, Yan, Zarov, Zhang, Fan,
  Kambadur, Narang, Rodriguez, Stojnic, Edunov, and
  Scialom}]{DBLP:journals/corr/abs-2307-09288}
Hugo Touvron, Louis Martin, Kevin Stone, Peter Albert, Amjad Almahairi, Yasmine
  Babaei, Nikolay Bashlykov, Soumya Batra, Prajjwal Bhargava, Shruti Bhosale,
  Dan Bikel, Lukas Blecher, Cristian Canton{-}Ferrer, Moya Chen, Guillem
  Cucurull, David Esiobu, Jude Fernandes, Jeremy Fu, Wenyin Fu, Brian Fuller,
  Cynthia Gao, Vedanuj Goswami, Naman Goyal, Anthony Hartshorn, Saghar
  Hosseini, Rui Hou, Hakan Inan, Marcin Kardas, Viktor Kerkez, Madian Khabsa,
  Isabel Kloumann, Artem Korenev, Punit~Singh Koura, Marie{-}Anne Lachaux,
  Thibaut Lavril, Jenya Lee, Diana Liskovich, Yinghai Lu, Yuning Mao, Xavier
  Martinet, Todor Mihaylov, Pushkar Mishra, Igor Molybog, Yixin Nie, Andrew
  Poulton, Jeremy Reizenstein, Rashi Rungta, Kalyan Saladi, Alan Schelten, Ruan
  Silva, Eric~Michael Smith, Ranjan Subramanian, Xiaoqing~Ellen Tan, Binh Tang,
  Ross Taylor, Adina Williams, Jian~Xiang Kuan, Puxin Xu, Zheng Yan, Iliyan
  Zarov, Yuchen Zhang, Angela Fan, Melanie Kambadur, Sharan Narang,
  Aur{\'{e}}lien Rodriguez, Robert Stojnic, Sergey Edunov, and Thomas Scialom.
  2023.
\newblock \href {https://doi.org/10.48550/arXiv.2307.09288} {Llama 2: Open
  foundation and fine-tuned chat models}.
\newblock \emph{CoRR}, abs/2307.09288.

\bibitem[{Wang et~al.(2021)Wang, Wang, Joty, and
  Hoi}]{DBLP:conf/emnlp/0034WJH21}
Yue Wang, Weishi Wang, Shafiq~R. Joty, and Steven C.~H. Hoi. 2021.
\newblock \href {https://doi.org/10.18653/v1/2021.emnlp-main.685} {Codet5:
  Identifier-aware unified pre-trained encoder-decoder models for code
  understanding and generation}.
\newblock In \emph{Proceedings of the 2021 Conference on Empirical Methods in
  Natural Language Processing, {EMNLP} 2021, Virtual Event / Punta Cana,
  Dominican Republic, 7-11 November, 2021}, pages 8696--8708. Association for
  Computational Linguistics.

\bibitem[{Wei et~al.(2022)Wei, Bosma, Zhao, Guu, Yu, Lester, Du, Dai, and
  Le}]{DBLP:conf/iclr/WeiBZGYLDDL22}
Jason Wei, Maarten Bosma, Vincent~Y. Zhao, Kelvin Guu, Adams~Wei Yu, Brian
  Lester, Nan Du, Andrew~M. Dai, and Quoc~V. Le. 2022.
\newblock \href {https://openreview.net/forum?id=gEZrGCozdqR} {Finetuned
  language models are zero-shot learners}.
\newblock In \emph{The Tenth International Conference on Learning
  Representations, {ICLR} 2022, Virtual Event, April 25-29, 2022}.
  OpenReview.net.

\bibitem[{Wu et~al.(2023)Wu, Irsoy, Lu, Dabravolski, Dredze, Gehrmann,
  Kambadur, Rosenberg, and Mann}]{DBLP:journals/corr/abs-2303-17564}
Shijie Wu, Ozan Irsoy, Steven Lu, Vadim Dabravolski, Mark Dredze, Sebastian
  Gehrmann, Prabhanjan Kambadur, David~S. Rosenberg, and Gideon Mann. 2023.
\newblock \href {https://doi.org/10.48550/arXiv.2303.17564} {Bloomberggpt: {A}
  large language model for finance}.
\newblock \emph{CoRR}, abs/2303.17564.

\end{thebibliography}
\bibliographystyle{acl_natbib}

\clearpage

\appendix

\section{Appendix}
\label{sec:appendix}

\subsection{Task Definitions}

Definitions of the tasks are provided in Table \ref{task_defintion}.

\label{appendix:task-definitions}

\begin{table*}[ht]
\footnotesize
\centering
\begin{tabular}{l|l}
\textbf{Task}                & \textbf{Definition}                                \\
\midrule
Call Reason                  & What is the primary call intent                    \\
Call Summarization           & Summarize the dialog                               \\
Segmented Call Summarization & Summarize a segmented portion of the dialog        \\
Confirmed Next Steps         & List the confirmed next steps if any in the dialog \\
Question-Answering (QA) & Answer the question based on context present in the dialog               \\
Entity Extraction       & List the entities present in the dialog             \\
Topic Segmentation           & Segment the dialog into coherent topics            \\
Text Rewriting (QA)     & Rewrite a given piece of text in a fluent and grammatically correct form \\
\end{tabular}
\caption{Definitions of tasks considered in the internally curated contact-center dataset}
\label{task_defintion}
\end{table*}


\end{document}